\documentclass[10pt,twocolumn,letterpaper]{article}

\usepackage{cvpr}              
\usepackage[accsupp]{axessibility}  

\usepackage{tabularx}  
\usepackage{array}     
\newcolumntype{C}{>{\centering\arraybackslash}X}

\definecolor{cvprblue}{rgb}{0.21,0.49,0.74}
\usepackage[pagebackref,breaklinks,colorlinks,allcolors=cvprblue]{hyperref}
\usepackage{amsmath, amssymb, graphicx, booktabs}
\usepackage[utf8]{inputenc}
\usepackage{algorithm, algpseudocode}
\usepackage{makecell, tabularx}
\usepackage{threeparttable}
\usepackage{tabularx}
\usepackage{multirow}
\newcolumntype{Y}{>{\raggedright\arraybackslash}X}
\usepackage{siunitx}
\newcommand{\ve}[1]{\mathbf{#1}} 

\newcommand{\ftxt}{f_{\text{txt}}}

\newcommand{\ehat}{\hat{\ve{e}}}
\newcommand{\MSE}[2]{\operatorname{MSE}\!\left(#1,\,#2\right)}
\def\paperID{*****} 
\def\confName{CVPR}
\def\confYear{2026}

\title{Similarity-as-Evidence: Calibrating Overconfident VLMs for Interpretable and Label-Efficient Medical Active Learning}

\author{
Zhuofan Xie$^{1\dagger}$ \quad
Zishan Lin$^{1\dagger}$ \quad
Jinliang Lin$^{2}$ \quad
Jie Qi$^{1}$ \quad
Shaohua Hong$^{2*}$ \quad
Shuo Li$^{3}$ \\
$^1$ School of Electronic Technology and Engineering, Xiamen University, Xiamen, China \\
$^2$ School of Informatics, Xiamen University, Xiamen, China \\
$^3$ School of Engineering, Case Western Reserve University, Cleveland, USA \\
\small{$^\dagger$ Equal contribution. $^*$ Corresponding author.}%
}
\begin{document}
\maketitle
\begin{abstract}
Active Learning (AL) reduces annotation costs in medical imaging by selecting only the most informative samples for labeling, but suffers from cold-start when labeled data are scarce. Vision-Language Models (VLMs) address the cold-start problem via zero-shot predictions, yet their temperature-scaled softmax outputs treat text-image similarities as deterministic scores while ignoring inherent uncertainty, leading to overconfidence. This overconfidence misleads sample selection, wasting annotation budgets on uninformative cases. To overcome these limitations, the Similarity-as-Evidence (SaE) framework calibrates text–image similarities by introducing a Similarity Evidence Head (SEH), which reinterprets the similarity vector as evidence and parameterizes a Dirichlet distribution over labels. In contrast to a standard softmax that enforces confident predictions even under weak signals, the Dirichlet formulation explicitly quantifies lack of evidence (vacuity) and conflicting evidence (dissonance), thereby mitigating overconfidence caused by rigid softmax normalization. Building on this, SaE employs a dual-factor acquisition strategy: high-vacuity samples (e.g., rare diseases) are prioritized in early rounds to ensure coverage, while high-dissonance samples (e.g., ambiguous diagnoses) are prioritized later to refine boundaries, providing clinically interpretable selection rationales. Experiments on ten public medical imaging datasets with a 20\% label budget show that SaE attains state-of-the-art macro-averaged accuracy of 82.57\%. On the representative BTMRI dataset, SaE also achieves superior calibration, with a negative log-likelihood (NLL) of 0.425.

\end{abstract}

\section{Introduction}
\label{sec:intro}
\begin{figure}[t]
    \centering
    \includegraphics[width=\linewidth]{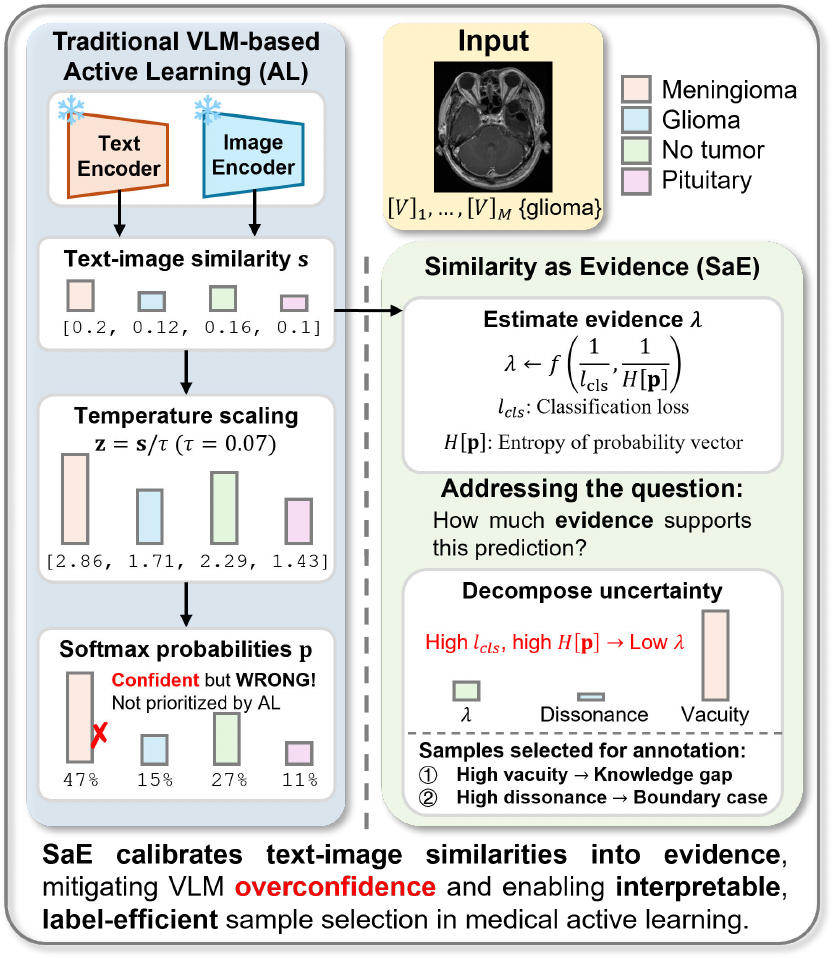}
    \caption{SaE addresses overconfidence in VLM-based AL by reframing text-image similarities as Dirichlet evidence. \textbf{Left:} Traditional VLM-based methods rely on temperature-scaled softmax, producing overconfident probabilities that mislead sample selection. \textbf{Right:} SaE calibrates similarities into evidence, enabling decomposition into vacuity (knowledge gaps) and dissonance (decision conflicts) for interpretable sample acquisition.}
    \label{fig:Fig1}
\end{figure}

Deep learning has achieved remarkable success in medical image analysis~\cite{1_DLinMIA}, yet its clinical deployment remains constrained by the scarcity of expert annotations, as the labeling process is often impeded by time, cost, and privacy regulations~\cite{2_labelSecure, 4_AL_mia_seg}. 
Active Learning (AL) addresses this challenge by iteratively selecting the most informative unlabeled samples for annotation, thereby maximizing model performance under limited labeling budgets~\cite{3_AL}. However, conventional AL methods face a critical cold-start problem: with only a handful of initial labeled samples (e.g., 1–3 per class), the model produces unreliable predictions that yield poor sample selection in early rounds, leading to inefficient use of annotation resources~\cite{cold_start-v227-chen24a, wang2024survey_deep_active_learning_mia}.

Vision--Language Models (VLMs) provide a promising means to address the cold-start by leveraging pre-trained text-image alignment for zero-shot and few-shot prediction~\cite{safaei2024_al_vlms, vlmfh}. By freezing the VLM encoders, even a shallow classifier can form relatively stable decision boundaries with minimal labeled data~\cite{2022CLIP_fewshot, CLIP_few_shot_2023_ICCV, CLIP_2024_CVPR}. Yet this strategy introduces a new challenge: \textbf{VLMs are inherently overconfident}. As illustrated in Fig.~\ref{fig:Fig1}, VLMs compute cosine similarities between image and text embeddings and convert them to probabilities via temperature-scaled softmax, effectively treating geometric proximity as certainty. This reliance on a contrastive formulation and its rigid softmax normalization induces severe miscalibration~\cite{vlm_overconfidence_2025_CVPR, overconfidence_vlm}, where the model assigns high probabilities to incorrect predictions. In AL, such overconfidence misleads the acquisition function: the model confidently selects samples it already believes it understands, rather than those that would maximally improve performance, wasting valuable annotation budget. Existing calibration methods, such as post-hoc temperature scaling~\cite{pmlr-v70-guo17a, NEURIPS2021_8420d359}, provide only global adjustments and do not explain why a prediction is uncertain~\cite{ZHAO2025103677}. 

In clinical settings, interpretability is critical. However, most AL strategies in medical imaging instead rely on scalar uncertainty scores such as predictive entropy or margin~\cite{kim2024margin_brain_tumor_segmentation_jdi, huang2024bald_entropy_jendo}. These uncertainty scores rank samples by the magnitude of uncertainty but they do not reveal whether uncertainty arises from missing knowledge or from conflicting hypotheses~\cite{budd2021survey_medical_imaging_al}. In clinical workflows, experts must understand why a case is selected for annotation and whether it reflects an unseen phenotype or an ambiguous decision boundary~\cite{wang2024survey_deep_active_learning_mia}. Subjective Logic (SL)~\cite{josang2016subjective, evidential_dl_neurips18} provides a formalism for this requirement by representing predictions as evidence that parameterizes a Dirichlet distribution and by decomposing uncertainty into vacuity, which measures knowledge gaps, and dissonance, which measures conflicts between competing labels. However, adapting this evidential view to VLM-based AL remains largely unexplored.

We propose \textbf{Similarity-as-Evidence (SaE)}, a framework that recalibrates VLM outputs into interpretable evidential uncertainty for medical AL. As shown in Fig.~\ref{fig:Fig2}, SaE introduces a Similarity Evidence Head (SEH) that maps raw text-image similarities to Dirichlet evidence parameters, quantifying how much evidence the VLM has for each prediction. Trained with a dual-objective loss that balances classification performance and uncertainty calibration, SEH produces calibrated evidence that mitigates overconfidence. We then decompose this evidence into vacuity and dissonance via SL, and design a dual-factor acquisition strategy:
\begin{enumerate}[(i)]
    \item \textbf{High-vacuity samples} (e.g., a pneumothorax case when the model has seen mostly normal lungs) are prioritized in early rounds to expand coverage of under-represented phenotypes.
    \item \textbf{High-dissonance samples} (e.g., a chest X-ray with features consistent with both pneumonia and pulmonary edema) are prioritized in later rounds to refine ambiguous decision boundaries.
\end{enumerate}
This adaptive strategy aligns sample selection with clinical reasoning and provides clinically interpretable rationales for annotation requests.


Our contributions are summarized as follows:
\begin{enumerate}
\item Similarity-as-Evidence (SaE) is the first framework to address VLM overconfidence in medical AL by mapping raw similarities to Dirichlet evidence, enabling calibrated and interpretable uncertainty quantification.
\item SaE introduces a dual-factor acquisition strategy that decomposes uncertainty into vacuity (knowledge gaps) and dissonance (decision conflicts), supporting adaptive sample selection with clinically interpretable rationales across AL rounds.
\item Extensive experiments across ten diverse medical datasets demonstrate SaE's superiority by achieving state-of-the-art (SOTA) label efficiency and offering clinically interpretable uncertainty.
\end{enumerate}

\section{Related Work}

\label{sec:relatedwork}

\subsection{Active Learning for Medical Image Analysis}

Traditional pool-based AL in medical imaging mainly relies on statistical uncertainty or diversity, but suffers from severe cold-start issues when labeled data are scarce \cite{budd2021survey_medical_imaging_al,wang2024survey_deep_active_learning_mia}.
Uncertainty sampling variants such as Least-Confidence \cite{belo2023least_confidence_chestxray_spie}, Margin \cite{kim2024margin_brain_tumor_segmentation_jdi}, and Entropy \cite{huang2024bald_entropy_jendo} efficiently target decision boundaries, yet are highly sensitive to artifacts and class imbalance \cite{schmidt2024focal_mia,ma2024selective_uncertainty_mia}, often re-selecting low-quality or intrinsically ambiguous scans rather than clinically informative ones \cite{budd2021survey_medical_imaging_al,wang2024survey_deep_active_learning_mia}.
Diversity-driven and hybrid methods \cite{li2023halia_mia,4_AL_mia_seg,kang2024wise_mia} reduce redundancy via clustering \cite{qiu2024caption_cluster_wsi_miccai}, core-sets \cite{vepa2024coreset_neurips}, or submodular selection \cite{kothawade2022submodular_mi_htl}, but incur high computational overhead in high-dimensional feature spaces and require careful fusion with uncertainty \cite{yang2017suggestive,kirsch2019batchbald}.
Bayesian AL (e.g., BALD \cite{huang2024bald_entropy_jendo}) offers an information-theoretic criterion via mutual information but typically relies on posterior sampling such as MC Dropout \cite{sadafi2023wsi_mcdropout_isbi}, which is costly and still does not explain \emph{why} predictions disagree \cite{kirsch2019batchbald}.
Overall, existing AL strategies lack semantic awareness and fail to articulate why a case is informative in clinical terms, motivating the integration of VLMs to inject rich, semantically grounded evidence into selection.

\subsection{Vision-Language Models in Medical Diagnostics}

\label{sec:vlm_related}

VLMs align images and text in a shared embedding space \cite{MediCLIP,BiomedCLIP,zhang2023kad}, enabling zero- and few-shot classification of radiologic findings from textual prompts.
In chest X-rays, such zero-shot methods already achieve competitive performance across multiple datasets with minimal labels \cite{tiu2022chexzero,jang2024zeroshoot_cxr}.
However, recent work shows that VLMs are often severely miscalibrated and overconfident, assigning high similarity scores to incorrect classes, especially under distribution shift or naive fine-tuning \cite{levine2023vlm_calibration,yoon2024ctpt,vlm_overconfidence_2025_CVPR,yue2024_calib_vlm}.
Analyses of CLIP-like models further reveal weak concept binding, high background sensitivity, and limited faithfulness of similarity scores \cite{li2023clip_explainability,lewis2024does_clip_bind,splice2024}, indicating that raw similarity is not a causal explanation for predictions.
Consequently, naively using VLM scores for AL risks populating the labeled set with confidently wrong samples; a model is needed to convert similarities into calibrated, clinically meaningful evidence.

\subsection{Uncertainty Quantification in Deep Learning}

Reliable uncertainty quantification is central to robust AL. Classical approaches such as MC Dropout and deep ensembles \cite{gal2016dropout,Bargagna2023BayesianCN,lakshminarayanan2017deepensembles,Gu_Revisiting_MICCAI2024} provide strong predictive uncertainty but require multiple forward passes or several models, making them expensive for large-scale medical pipelines and iterative AL cycles \cite{uncertaintyQuantifyComputational, huang2024bald_entropy_jendo}.
Evidential Deep Learning (EDL) offers a more efficient alternative by modeling predictions as a Dirichlet distribution whose concentration parameters encode evidence under SL \cite{gao2024_edl_survey,hu2021_muanet,josang2016subjective}.
This representation enables decomposing uncertainty into more interpretable quantities, and has been explored for medical AL, for example in FedEvi \cite{Chen_2024_CVPR_EDL_AL} and vacuity-aware acquisition criteria \cite{zhou2024_deal_uas}.
Yet standard EDL typically derives evidence from direct transformations of class-probability vectors, which can behave brittlely in early AL rounds and under distribution shift or class imbalance \cite{gao2024_edl_survey,hu2021_muanet,danruo2021_iedl}.
Moreover, evidential losses are highly sensitive to regularization; improper choices inflate spurious epistemic uncertainty and bias acquisition \cite{ulmer2023_priorposterior,mixture2024_edl_unify}.
We therefore propose to anchor evidential quantities in the rich semantic space of VLMs, converting image–text similarities into calibrated evidence to stabilize cold-start behavior and yield clinically interpretable uncertainty signals \cite{safaei2024_al_vlms,yue2024_calib_vlm,baumann2024_bayesvlm}.

\subsection{Active Learning with Vision-Language Models}
Recent work has begun to couple VLMs with AL, primarily using them as feature extractors or for zero-shot pseudo-labeling \cite{safaei2024_al_vlms,PCB}.
Other approaches attempt to incorporate VLM-based uncertainty, e.g., by calibrating VLM entropy and combining it with neighbor-aware uncertainty \cite{safaei2024_al_vlms} or using VLM priors to guide prompt selection and class coverage \cite{PCB}.
However, these methods largely reduce VLM uncertainty to a single scalar derived from softmax probabilities, thereby inheriting overconfidence issues and lacking a principled decomposition of uncertainty sources.
In particular, they cannot distinguish lack of knowledge from conflicting evidence and thus limit the clinically meaningful interpretation of query decisions.
\section{Method}
\label{sec:method}
\subsection{Preliminaries}
\label{sec:preliminaries}
\begin{figure*}[ht]
    \centering
    \includegraphics[width=0.85\linewidth]{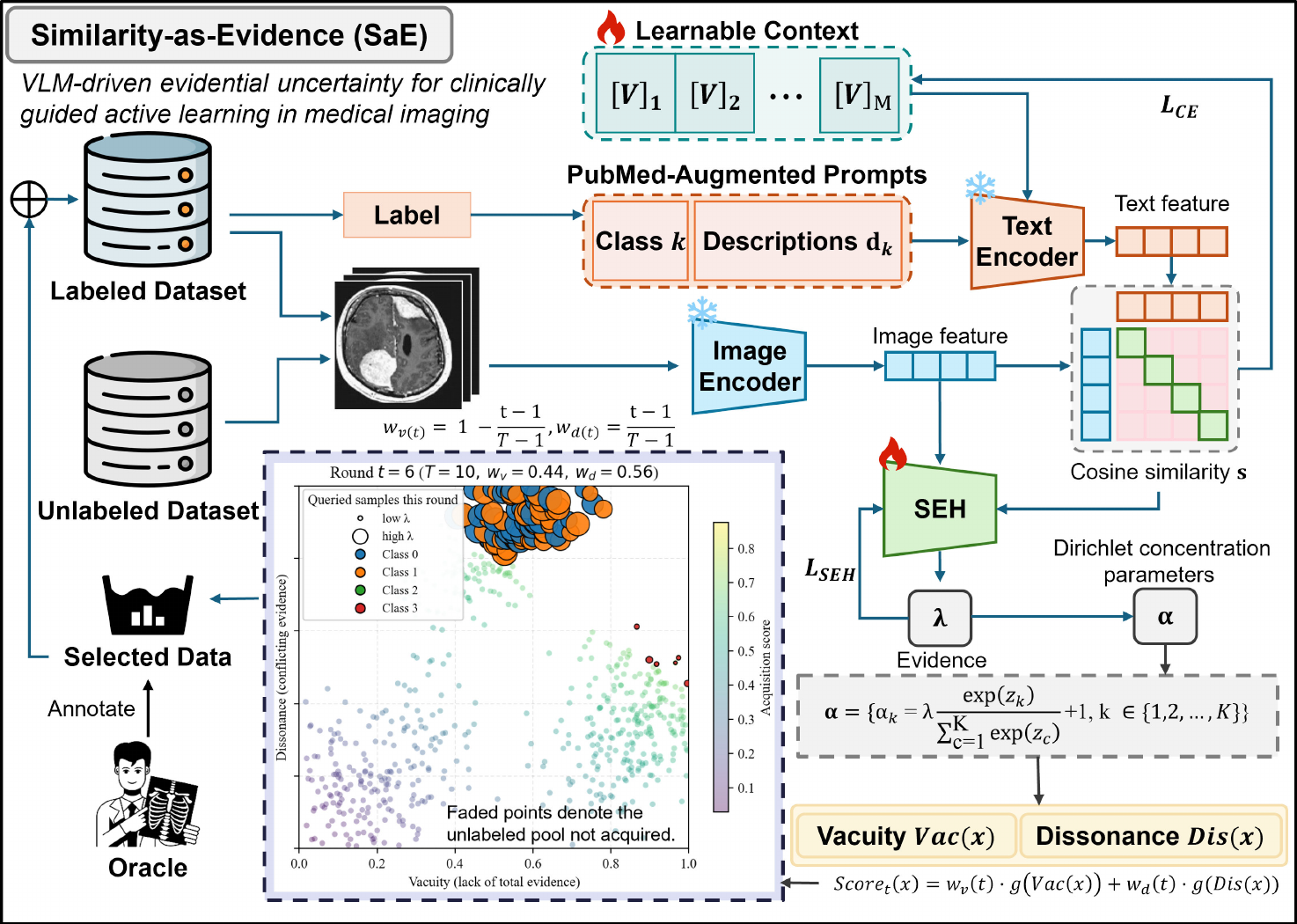}
    \caption{Overview of the proposed SaE framework. A frozen VLM encodes images and PubMed-augmented class prompts to produce text-image similarities, which are mapped by a trainable SEH into Dirichlet evidence. We decompose this evidence into vacuity and dissonance, and use them to score unlabeled samples for AL. The inset shows one acquisition round: high-vacuity cases are prioritized early to cover unseen phenotypes, and high-dissonance boundary disputes are prioritized later for expert annotation.}
    \label{fig:Fig2}
\end{figure*}
\paragraph{Active Learning.} We consider the standard pool-based AL setting, which starts with a large unlabeled pool $\mathcal{D}_U$ and a small initial labeled seed set $\mathcal{D}_L = \{(x_i, y_i)\}_{i=1}^{N}$. At each round $t$, an acquisition function selects a query batch $\mathcal{D}_Q^t \subset \mathcal{D}_U$ of the most informative samples based on the current model. These samples are annotated by an oracle (e.g., a radiologist) and moved to $\mathcal{D}_L$: $\mathcal{D}_L \leftarrow \mathcal{D}_L \cup \mathcal{D}_Q^t$, $\mathcal{D}_U \leftarrow \mathcal{D}_U \setminus \mathcal{D}_Q^t$. The model is then retrained on the updated $\mathcal{D}_L$. This process repeats for $T$ rounds or until a labeling budget is exhausted.

\paragraph{Evidential Deep Learning (EDL).} 
EDL extends standard deep classification by modeling a second-order probability distribution over class assignments, enabling the separation of aleatoric and epistemic uncertainty. Instead of predicting a single probability vector $\ve{p}$, EDL places a Dirichlet distribution $\text{Dir}(\ve{p}|\boldsymbol{\alpha})$ over the categorical distribution. The concentration parameters $\boldsymbol{\alpha} = [\alpha_1, \dots, \alpha_K]$ for $K$ classes are predicted by a neural network, where $\alpha_k = e_k + 1$ and $e_k \ge 0$ represents the collected evidence for class $k$. The total evidence is $S = \sum_{k=1}^K \alpha_k$, and the expected class probability is $\hat{p}_k = \alpha_k / S$.

\paragraph{Vision-Language Models (VLMs).} 
A VLM (e.g., CLIP~\cite{CLIP} or its medical variants such as MedCLIP~\cite{MedCLIP}, BiomedCLIP~\cite{BiomedCLIP}) comprises an image encoder $f_{\mathrm{img}}$ and a text encoder $f_{\mathrm{txt}}$, aligned via contrastive pre-training. Let $\hat{\mathbf{e}}_{\mathrm{img}} = f_{\mathrm{img}}(x) / \|f_{\mathrm{img}}(x)\|_2$ be the normalized embedding for an image $x$. For zero-shot classification over $K$ classes $\{c_k\}_{k=1}^{K}$, hand-crafted prompts (e.g., ``a photo of a $c_k$'') are encoded to obtain class prototypes $\{\hat{\mathbf{e}}^{\,k}_{\mathrm{txt}}\}_{k=1}^{K}$. The prediction is based on the cosine similarities $s_k = \langle \hat{\mathbf{e}}_{\mathrm{img}}, \hat{\mathbf{e}}^{\,k}_{\mathrm{txt}}\rangle$. Standard VLMs convert these similarities to probabilities via a temperature-scaled softmax: $\pi_k = \frac{\exp(s_k/\tau)}{\sum_{j=1}^{K}\exp(s_j/\tau)}$. In this work, we typically keep the heavy image encoder $f_{\mathrm{img}}$ frozen to leverage its robust pre-trained features.


\paragraph{Prompt Learning for Adaptation.} 
To adapt general VLMs to specific domains without full fine-tuning, prompt learning (e.g., CoOp~\cite{CoOp}) optimizes $M$ learnable context vectors $\{V_m\}_{m=1}^{M}$ while keeping the encoders frozen. A class prompt becomes $\mathcal{P}_k = [V_1, \dots, V_M, \mathbf{c}_k]$, where $\mathbf{c}_k$ is the fixed word embedding of the class name $c_k$. These learned prompts generate task-specific text prototypes $\hat{\mathbf{e}}^{\,k}_{\mathrm{txt}}$, which are optimized via cross-entropy loss on $\mathcal{D}_L$. This technique is crucial in medical settings to align the VLM's semantic space with specialized clinical concepts using limited data.



\subsection{Similarity-as-Evidence (SaE)}
\label{sec:sae}
The SaE framework addresses VLM miscalibration and enhances AL interpretability through three synergistic components (Fig.~\ref{fig:Fig2}). First, PubMed-augmented prompts enrich the domain-specific semantics for the VLM. Second, a Similarity Evidence Head (SEH) parameterizes a Dirichlet distribution directly from the VLM's similarity vector. Third, a dual-factor AL strategy decomposes the resulting evidential uncertainty (vacuity and dissonance) to guide sample selection. We detail each component in the following sections.

\subsubsection{PubMed-Augmented Prompts}
\label{sec:pubmend_aug}
Bridging the gap between the general knowledge of VLMs and the specific semantics of medical imaging requires enriched textual representations. We therefore augment class prompts with specialized knowledge retrieved from PubMed~\cite{pubmed}. For each class $k$, we retrieve $\delta_k$ descriptive sentences $\{d_{k,i}\}_{i=1}^{\delta_k}$ from PubMed. Each description is inserted into a prompt template and encoded by the frozen text encoder $\ftxt$. The resulting embeddings are L2-normalized and averaged to form a robust class-prototype embedding $\bar{\ehat}_{\text{txt}}^{\,k}$:

\begin{equation}
\begin{aligned}
\bar{\ehat}_{\text{txt}}^{\,k}
&= \frac{1}{\delta_k}\sum_{i=1}^{\delta_k}\frac{\ftxt(d_{k,i})}{\|\ftxt(d_{k,i})\|_2}.
\end{aligned}
\end{equation}
The similarity score for class $k$ is the cosine similarity between the normalized image embedding $\ehat_{\text{img}}$ and this prototype: $s_k=\langle \ehat_{\text{img}},\bar{\ehat}_{\text{txt}}^{,k} \rangle $. This process produces a semantically rich similarity vector $\ve{s} = [s_1, \dots, s_K]$ that serves as the input to our evidence model. See supplementary Sec.~\ref{sec:prompts} for details.

\subsubsection{Similarity Evidence Head (SEH)}
\label{sec:seh}
While a VLM's raw similarity vector $\mathbf{s}$ is often miscalibrated for direct decision-making, it contains a rich prior regarding the model's intrinsic uncertainty. Standard methods discard this information by forcing a rigid softmax normalization. The SEH, detailed in supplementary Sec.~\ref{sec:SEH-arch}, harnesses this prior by reframing the similarity vector as the relative allocation ratios for a total evidence budget.

Specifically, given the frozen VLM image embedding $\mathbf{x} = f_{\mathrm{img}}(x)$ and its corresponding class-similarity vector $\mathbf{s}$, SEH estimates a scalar strictly positive evidence strength $\lambda$. This parameter $\lambda$ is trained to jointly reflect sample difficulty and intrinsic VLM confidence. SEH adopts a dual-branch MLP architecture with batch normalization (BN), ReLU, and dropout (rate $0.1$). The feature branch encodes $\mathbf{x}$ into $z_f$ via a two-layer MLP, while the similarity branch maps $\mathbf{s}$ to $z_s$. The concatenated representation $[z_f; z_s]$ is fused by a shallow MLP followed by a softplus activation to enforce positivity:
\begin{equation}
\begin{aligned}
z_f &= f_{\text{img}}^{\text{SEH}}(\mathbf{x}), \quad
z_s = f_{\text{sim}}^{\text{SEH}}(\mathbf{s}), \\
\tilde{\lambda} &= f_{\text{fuse}}^{\text{SEH}}([z_f; z_s]), \quad
\lambda = \operatorname{softplus}(\tilde{\lambda}).
\end{aligned}
\end{equation}
where the softplus activation is defined as $\operatorname{softplus}(u) = \log\bigl(1 + e^{u}\bigr)$.

\paragraph{SEH Loss Function.}
We design a dual-objective loss function $\mathcal{L}_{\text{SEH}}$ to train the SEH for producing meaningful evidence strength $\lambda$. For a batch of $N_B$ samples, let $l_{\text{cls},i} = -\log p_{i, y_i}$ be the cross-entropy loss for sample $i$ using its ground truth $y_i$. The total SEH loss is:
\begin{equation}
\begin{aligned}
&\mathcal{L}_{\text{SEH}} = \MSE{\frac{1}{\lambda_i + \epsilon}}{\,l_{\text{cls},i}} + \beta\,\MSE{\lambda_i}{\,\frac{1}{H[\mathbf{p}_i]+\epsilon}},\\
&\operatorname{MSE}(a_i,b_i) \triangleq \frac{1}{N_B}\sum_{i=1}^{N_B} (a_i-b_i)^2.
\label{eq:seh-loss}
\end{aligned}
\end{equation}
Here, $H[\mathbf{p}_i]=\text{softmax}(\mathbf{s}_i/\tau_{\mathrm{f}})$ is the entropy of the probability vector derived from the frozen VLM with fixed temperature $\tau_{\mathrm{f}}$, and $\epsilon$ is a small constant for numerical stability. Crucially, $H[\mathbf{p}_i]$ serves as a detached target, ensuring that gradients do not propagate back into the frozen VLM. The coefficient $\beta$ (default $0.5$) balances the two terms. The first term aligns inverse evidence with observed classification difficulty ($1/\lambda \uparrow$ for hard samples with high $l_{\text{cls}}$). The second term enforces consistency with the VLM's intrinsic certainty (low entropy $\Rightarrow$ high $\lambda$). This synergistic loss trains SEH to produce a calibrated $\lambda$ that reflects both empirical difficulty and latent VLM confidence.

\begin{algorithm}[t]
\caption{Dual-Factor Active Learning Strategy}
\label{alg:sae_al}
\begin{algorithmic}[1]
\Require Labeled set $\mathcal{D}_{L}$, unlabeled pool $\mathcal{D}_{U}$, SEH model $M_{\text{SEH}}$,  per-round budget ratio $\rho_t$, number of rounds $T$, number of classes $K$.
\For{$t=1$ to $T$}
    \If{$\mathcal{D}_{U}=0$} \State \textbf{break} \EndIf
    \State $N_t \gets \min\!\big(\,\mathcal{D}_{U},\;\max(1,\; \rho_t\times\mathcal{D}_{U})\big)$
    \State Initialize empty list $\text{Scores} \gets [\;]$ \Comment{store pairs $(x,\text{Score}_t(x))$}
    \For{each sample $x \in \mathcal{D}_{U}$}
        \State Obtain similarity $\ve{s}(x)$ from frozen VLM
        \State Predict evidence strength $\lambda(x) \gets M_{\text{SEH}}(x, \ve{s}(x))$
        \State Compute $\alpha_k(x)$ via Eq.~\eqref{eq:evi-param-en} 
        \State Compute $\mathrm{Vac}(x),\;\mathrm{Dis}(x)$ from $\{\alpha_k(x)\}_{k=1}^{K}$ via Eq.~\eqref{eq:vacuity},~\eqref{eq:dissonance}
        \State $\text{Score}_t(x) \gets$ combine $\mathrm{Vac}(x)$ and $\mathrm{Dis}(x)$ using Eq.~\eqref{eq:al_score} with schedule in Eq.~\eqref{eq:schedule}
        \State Append $(x,\text{Score}_t(x))$ to $\text{Scores}$
    \EndFor
    \State $\mathcal{D}_Q^t$ $\gets$ indices of the top-$N_t$ elements in $\text{Scores}$ by descending $\text{Score}_t(x)$
    \State Annotate samples in $\mathcal{D}_Q^t$
    \State $\mathcal{D}_{L} \gets \mathcal{D}_{L} \cup \mathcal{D}_Q^t$; \quad $\mathcal{D}_{U} \gets \mathcal{D}_{U} \setminus \mathcal{D}_Q^t$
\EndFor
\State \textbf{return} Trained model $M_{\text{SEH}}$
\end{algorithmic}
\end{algorithm}
\subsubsection{Dual-Factor Active Learning Strategy}
\label{sec:al}
The SEH transforms the informative but miscalibrated similarity vector into a reliable source of evidence, providing the foundation for our dual-factor AL strategy.

\paragraph{Similarity-to-Evidence Mapping.}
For each sample $x$ in the unlabeled pool $\mathcal{D}_U$, we construct Dirichlet concentration parameters $\boldsymbol{\alpha}$ by scaling the base VLM probability distribution $\ve{p}(x)$ with the predicted evidence strength $\lambda(x)$:

\begin{equation}
    \alpha_k(x) = \lambda(x) \cdot p_k(x) + 1 = \lambda(x) \cdot \frac{\exp(s_k(x)/\tau)}{\sum_{c=1}^{K}\exp(s_c(x)/\tau)} + 1.
    \label{eq:evi-param-en}
\end{equation}

Note that we include the temperature $\tau$ here for generality, though it is often fixed in pre-trained VLMs. This Dirichlet posterior enables a principled decomposition of uncertainty into two clinically meaningful factors under SL \cite{josang2016subjective}:

\begin{itemize}
\item \textbf{Vacuity (lack of evidence):} defined as \begin{equation}
        \mathrm{Vac}(x) = K / \sum_{k=1}^K \alpha_k(x)
        \label{eq:vacuity}
    \end{equation} 
    where $S(x) = \sum_{k=1}^K \alpha_k(x)$ is the total evidence. High vacuity flags rare or currently unseen phenotypes where the model lacks sufficient total evidence to make a judgment.
\item \textbf{Dissonance (conflicting evidence):} measures the conflict among competing classes. We first compute belief masses $b_k(x)$, which are equivalent to the expected class probabilities $\hat{p}_k$ from the EDL preliminary: $b_k(x) = \alpha_k(x) -1/ S(x)$. Dissonance is then quantified as:
\begin{equation}
    \begin{split}
    \mathrm{Dis}(x)=\sum_{i=1}^K \frac{b_i(x) \times \sum_{j \neq i} \bigl[b_j(x) \cdot \mathrm{Bal}(b_i(x), b_j(x))\bigr]}{\sum_{j \neq i} b_j(x)}, \\
    \text{where} \quad \mathrm{Bal}(b_i,b_j)=1-\frac{|b_i-b_j|}{b_i+b_j}.
    \end{split}
    \label{eq:dissonance}
\end{equation}
\end{itemize}
\paragraph{Dual-Factor Score.}
The AL process proceeds in $T$ rounds until a total labeling budget ratio $\rho$ is exhausted. In each round $t$, we select a batch corresponding to a per-round ratio $\rho_t = \rho/T$. 
The acquisition score for each sample $x \in \mathcal{D}_{U}$ is a dynamic weighted combination of its normalized uncertainty factors:

\begin{equation}
\mathrm{Score}_t(x) = w_v(t) \cdot g(\mathrm{Vac}(x)) + w_d(t) \cdot g(\mathrm{Dis}(x)),
\label{eq:al_score}
\end{equation}
where $g(\cdot)$ is min-max normalization applied across $\mathcal{D}_{U}$. We employ a linear schedule: 
\begin{equation}
    w_v(t) = 1 - \frac{t-1}{T-1}, \quad w_d(t) = \frac{t-1}{T-1}
    \label{eq:schedule}
\end{equation}
This schedule gradually shifts the selection focus from prioritizing high-vacuity, unknown samples in early rounds to focusing on high-dissonance, hard-to-distinguish cases in later rounds. The overall process is summarized in Algorithm~\ref{alg:sae_al}.

\section{Experiments}
\label{sec:experiments}
\subsection{Experimental Setup}
\paragraph{Datasets.}
We evaluate our method on ten public medical imaging datasets covering nine organs (Table~\ref{tab:datasets}), using the same train-test splits as BiomedCoOp~\cite{BiomedCoOp}. The initial training set is treated as the unlabeled pool, from which samples are actively queried. Test sets are used only for final evaluation, and all reported numbers are Top-1 accuracy (\%).

\begin{table}[ht] 
\centering 
\small 
\caption{Evaluation is conducted on ten diverse medical datasets covering nine distinct organs. $K$ is the number of classes. Splits follow the official BiomedCoOp \cite{BiomedCoOp} benchmark.}
\label{tab:datasets} 
\begin{tabular}{l l l l} 
\toprule \textbf{Dataset} & \textbf{Organ(s)} & \textbf{$K$} & \textbf{\# train/val/test} \\ 
\midrule DermaMNIST \cite{DermaMNIST1,DermaMNIST2} & Skin & 7 & 
7007/1003/2005 \\ 
\midrule Kvasir \cite{Kvasir} & Colon & 8 & 
2000/800/1200 \\ 
\midrule RETINA \cite{RETINA1,RETINA2} & Retina & 4 & 
2108/841/1268 \\ 
\midrule LC25000 \cite{LC25000} & Lung, Colon & 5 & 
12500/5000/7500 \\ 
\midrule CHMNIST \cite{CHMNIST} & Colorectal & 8 & 
2496/1000/1504 \\ 
\midrule BTMRI \cite{BTMRI} & Brain & 4 & 
2854/1141/1717 \\ 
\midrule OCTMNIST \cite{OCTMNIST} & Retina & 4 & 
97477/10832/1000 \\ 
\midrule BUSI \cite{BUSI} & Breast & 3 & 
389/155/236 \\ \midrule COVID-QU-Ex \cite{COVIDQUEx} & Chest & 4 & 
10582/4232/6351 \\ 
\midrule KneeXray \cite{KneeXray} & Knee & 5 & 
5778/826/1656 \\ 
\bottomrule \end{tabular} 
\end{table}

\paragraph{Implementation details.}
Our AL pipeline runs for $T{=}5$ rounds. In each round $t$, we query $N_t = \rho_t \times \mathcal{D}_U$ samples, where $\rho_t = \rho / T$ and $\rho$ is the total budget ratio. Unless otherwise noted, we adopt BiomedCLIP (ViT‑B/16) as the vision–language backbone. The image encoder is frozen, and the text branch uses CoOp‑style learnable context with length $M{=}16$, and the prompt learner is enhanced as detailed in Sec.~\ref{sec:pubmend_aug}. This configuration forms our prompt‑learning baseline, denoted \textbf{MedCoOp}.
We train models using SGD (learning rate 0.002) with a cosine-annealing schedule for 100 epochs per round. The batch size is 32. All experiments are implemented in PyTorch~2.0.1 and run on a single NVIDIA RTX~4090 GPU.

\paragraph{Evaluation Metrics.}
Our primary evaluation metric is Top-1 accuracy (\%). To assess calibration, we also report Expected Calibration Error (ECE) and Negative Log Likelihood (NLL) without temperature scaling. For these calibration metrics, confidence is defined as the maximum class probability. 
\begin{table*}[ht]
\centering
\setlength{\tabcolsep}{4pt}
\caption{SaE achieves state-of-the-art AL performance, consistently outperforming all baselines on ten datasets. Results show mean accuracy (\%) and std. dev. across 5 seeds at a 20\% budget. BiomedCoOp is a few-shot reference.}
\label{tab:al_results}
\begin{tabular}{l c c c c c c c}
\toprule
\textbf{Dataset} & \textbf{Random} & \textbf{PCB}~\cite{PCB} & \makecell[c]{\textbf{MedCoOp}\\+Coreset~\cite{Coreset}} & \makecell[c]{\textbf{MedCoOp}\\+Entropy~\cite{Entropy}} & \makecell[c]{\textbf{MedCoOp}\\+BADGE\cite{BADGE}} & \textbf{BiomedCoOp}~\cite{BiomedCoOp} & \textbf{Ours (SaE)} \\
\midrule
DermaMNIST    & 69.42$\pm$0.5 & 71.07$\pm$0.6 & 74.11$\pm$0.7 & 74.56$\pm$0.5 & 75.46$\pm$0.6 & 62.59$\pm$1.8 & \textbf{80.21$\pm$0.4} \\
Kvasir        & 71.10$\pm$0.7 & 72.92$\pm$0.9 & 80.83$\pm$0.6 & 81.92$\pm$0.5 & 81.42$\pm$0.4 & 78.89$\pm$1.2 & \textbf{88.58$\pm$0.3} \\
RETINA        & 51.48$\pm$0.8 & 53.55$\pm$0.5 & 62.78$\pm$0.9 & 65.22$\pm$0.6 & 66.88$\pm$0.5 & 61.28$\pm$1.1 & \textbf{75.22$\pm$0.4} \\
LC25000       & 93.92$\pm$0.3 & 95.71$\pm$0.2 & 96.93$\pm$0.2 & 97.47$\pm$0.2 & 97.25$\pm$0.2 & 92.68$\pm$0.6 & \textbf{99.23$\pm$0.2} \\
CHMNIST       & 69.75$\pm$0.6 & 85.31$\pm$0.5 & 77.99$\pm$0.6 & 86.64$\pm$0.4 & 87.70$\pm$0.5 & 79.05$\pm$2.2 & \textbf{91.03$\pm$0.4} \\
BTMRI         & 83.40$\pm$0.4 & 85.50$\pm$0.4 & 86.26$\pm$0.3 & 89.92$\pm$0.3 & 89.57$\pm$0.4 & 83.30$\pm$1.3 & \textbf{93.46$\pm$0.2} \\
OCTMNIST      & 65.20$\pm$0.5 & 66.80$\pm$0.4 & 67.40$\pm$0.4 & 76.20$\pm$0.3 & 75.30$\pm$0.3 & 66.93$\pm$0.6 & \textbf{79.80$\pm$0.3} \\
BUSI          & 57.10$\pm$0.8 & 58.47$\pm$0.6 & 66.53$\pm$0.6 & 72.03$\pm$0.6 & 72.88$\pm$0.5 & 70.34$\pm$2.3 & \textbf{79.15$\pm$0.5} \\
COVID-QU-Ex   & 78.02$\pm$0.4 & 81.86$\pm$0.5 & 82.21$\pm$0.5 & 85.39$\pm$0.3 & 85.04$\pm$0.3 & 78.72$\pm$0.2 & \textbf{89.49$\pm$0.2} \\
KneeXray      & 40.66$\pm$0.7 & 42.87$\pm$0.6 & 43.36$\pm$0.5 & 44.57$\pm$0.5 & 46.01$\pm$0.5 & 39.69$\pm$1.8 & \textbf{49.50$\pm$0.4} \\
\midrule
\textbf{Macro Avg.} & 68.01 & 71.41 & 73.84 & 77.39 & 77.75 & 71.35 & \textbf{82.57} \\

\bottomrule
\end{tabular}
\end{table*}

\subsection{Comparison Study}
We compare SaE against Random and several PubMed-augmented baselines: PCB(AE)+BADGE~\cite{PCB} (hereafter PCB) and three MedCoOp-based learners~\cite{Coreset,BADGE,Entropy}, under a fixed labeling budget ($\rho = 0.2$).
All active methods share the same training schedule. 
BiomedCoOp under the ($K\times16$) setting is reported as a few-shot reference.
Table~\ref{tab:al_results} shows means over five seeds. SaE reaches a macro average of 82.57\%, while the strongest baseline MedCoOp+BADGE reaches 77.75\%. SaE ranks first on all ten datasets. We observe the largest margins on RETINA (+8.34\%). Significant gains are also seen on Kvasir (+6.66\%), BUSI (+6.27\%), as well as on large datasets such as LC25000 (+1.76\%) and OCTMNIST (+3.6\%). 
These results show consistent improvements beyond uncertainty and diversity heuristics. 

\begin{table}[ht]
\centering
\small
\caption{Ablation study confirms the SEH is the most critical component for SaE's performance gain. Macro average accuracy (\%) is reported over 10 datasets at a 20\% budget.}
\label{tab:ablation_study}
\begin{threeparttable}
\setlength{\tabcolsep}{6pt}
\renewcommand{\arraystretch}{1.1}

\begin{tabularx}{\columnwidth}{@{} Y S[table-format=2.2] @{}}
\toprule
\multicolumn{1}{@{}l}{\textbf{Variant}} & \multicolumn{1}{c@{}}{\textbf{Macro avg. (\%)}} \\
\midrule
Random                                         & 68.01 \\
\textit{+ Dual-factor score}\tnote{a}          & 73.35 \\
\textit{+ VLM (similarity for evidence)}\tnote{b} & 78.62 \\
\textbf{SaE: + SEH}\tnote{c}                   & \bfseries 82.57 \\
\bottomrule
\end{tabularx}

\begin{tablenotes}[flushleft]
\footnotesize
\item[a] Replace the sampler with a dual-factor score that uses vacuity and dissonance computed on classifier logits.
\item[b] Keep the dual-factor sampler, and replace classifier scores with BiomedCLIP similarity. SEH is not used.
\item[c] Add SEH to estimate calibrated evidence from similarity while keeping the sampler unchanged.
\end{tablenotes}
\end{threeparttable}
\end{table}

\subsection{Ablation Study}
We ablate three increments under the same AL protocol ($T=5$) and budget ($\rho=0.2$) over 10 datasets as shown in Table~\ref{tab:ablation_study}.
Replacing the Random sampler with a dual-factor score lifts the macro average from 68.01\% to 73.35\% (+5.34\%).
Using BiomedCLIP similarity to form Dirichlet evidence increases it to 78.62\% (+10.61\% \text{vs Random}).
Adding the SEH to calibrate the evidence reaches 82.57\%. This represents a 3.95\% gain over the previous step and a 14.56\% total gain over Random. 
These results show that the sampler provides a strong lift, the VLM adds a steady margin, and the SEH is necessary to reach optimal performance. Ablation studies on SEH loss components, hyperparameters, and acquisition schedule are in supplementary Sec.~\ref{sec:ablation}.

\begin{table}[h]
\centering
\small
\caption{SaE exhibits rapid early-round convergence, confirming its ability to mitigate the cold-start problem. Top-1 accuracy (\%) is shown after each round ($\rho{=}0.2$, $T{=}5$). The final column quantifies efficiency (ratio of $t=3$ to $t=5$ accuracy).}
\label{tab:round_efficiency}
\begin{threeparttable}
\setlength{\tabcolsep}{4pt}
\renewcommand{\arraystretch}{1.1}

\begin{tabularx}{\columnwidth}{@{}l
  S[table-format=2.2]
  S[table-format=2.2]
  S[table-format=2.2]
  S[table-format=2.2]
  S[table-format=2.2]
  !{\vrule width 0.5pt}
  S[table-format=2.2]@{}}
\toprule
\multicolumn{1}{@{}l}{\textbf{Dataset}} &
\multicolumn{1}{c@{}}{t=1} &
\multicolumn{1}{c@{}}{t=2} &
\multicolumn{1}{c@{}}{t=3} &
\multicolumn{1}{c@{}}{t=4} &
\multicolumn{1}{c@{}}{t=5} &
\multicolumn{1}{!{\vrule width 0.5pt}c@{}}{\textbf{$\frac{t{=}3}{t{=}5}$}} \\
\cmidrule(lr){2-6}\cmidrule(l){7-7}
DermaMNIST   & 72.07 & 76.11 & 77.56 & 78.46 & 80.21 & 96.70 \\
Kvasir       & 74.92 & 80.83 & 84.92 & 86.42 & 88.58 & 95.87 \\
RETINA       & 54.55 & 64.78 & 68.22 & 69.88 & 75.22 & 90.69 \\
LC25000      & 96.71 & 98.93 & 99.03 & 99.13 & 99.23 & 99.80 \\
CHMNIST      & 74.81 & 82.99 & 86.64 & 89.70 & 91.03 & 95.18 \\
BTMRI        & 86.50 & 88.26 & 92.92 & 93.02 & 93.46 & 99.42 \\
OCTMNIST     & 67.80 & 69.40 & 79.20 & 79.30 & 79.80 & 99.25 \\
BUSI         & 59.47 & 68.53 & 75.03 & 75.88 & 79.15 & 94.79 \\
COVID\text{-}QU\text{-}Ex & 82.86 & 84.21 & 88.39 & 88.49 & 89.49 & 98.77 \\
KneeXray     & 43.87 & 45.36 & 47.57 & 49.01 & 49.50 & 96.10 \\
\bottomrule
\end{tabularx}
\end{threeparttable}
\end{table}

\subsection{Analysis}
SaE is designed to convert VLM similarity into calibrated evidence and to spend labels where they deliver the largest gain. It targets two common failure modes in VLM-driven AL, namely cold-start instability and overconfident predictions. We therefore analyze early-round label efficiency and probability calibration.


\begin{figure}
    \centering
    \includegraphics[width=1\linewidth]{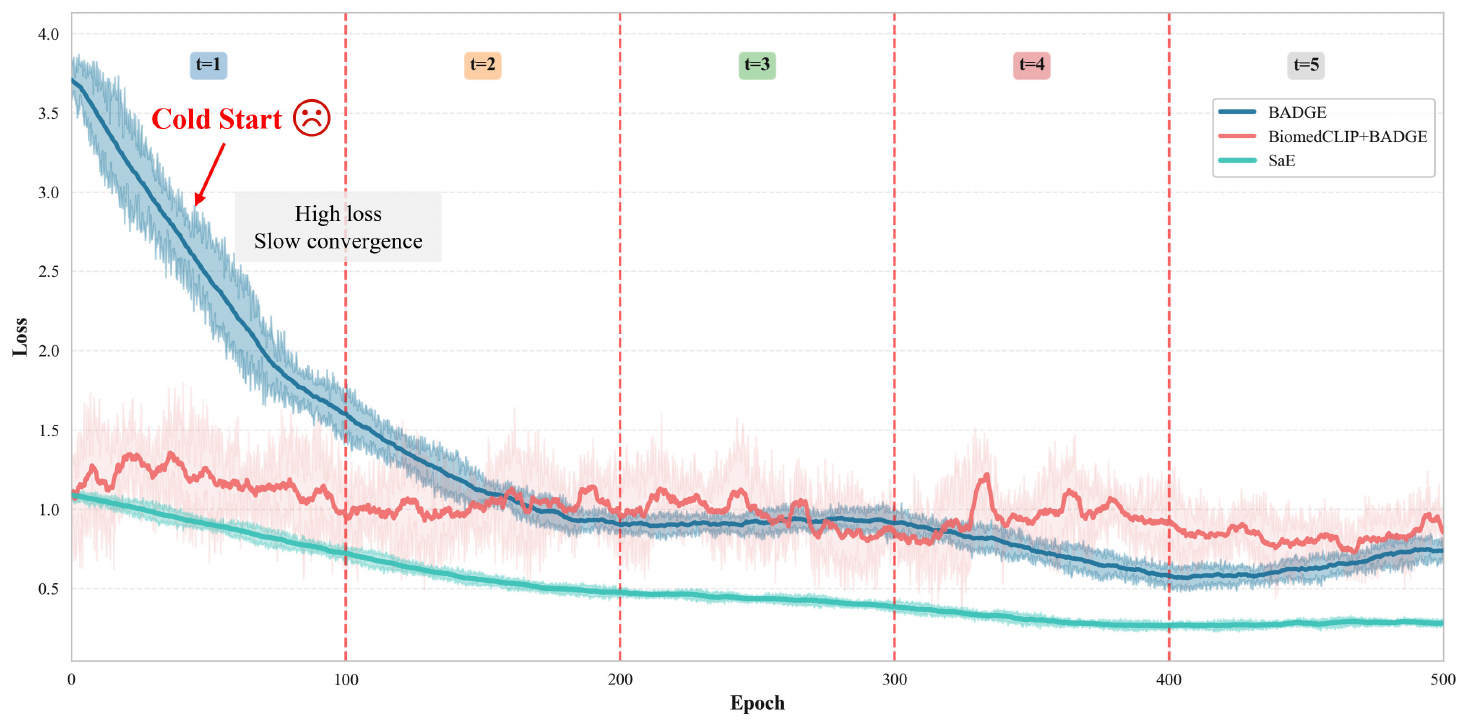}
    \caption{SaE achieves the lowest and most stable training loss across all AL rounds. Training trajectories on BTMRI ($\rho=0.2$) are compared. SaE avoids the high initial loss of BADGE and the volatility of BiomedCLIP+BADGE. Dashed red lines mark round transitions.}
    \label{fig:loss-trajectory}
\end{figure}

\begin{figure*}[t]
    \centering
    \includegraphics[width=0.8\linewidth]{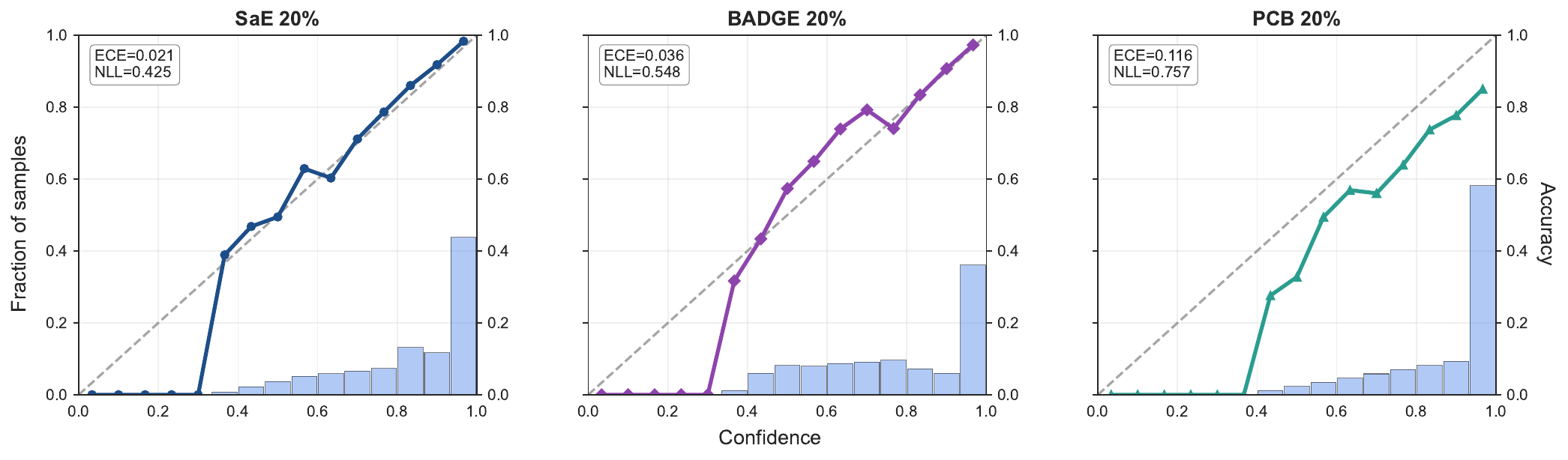}
    \caption{SaE achieves superior calibration and mitigates VLM overconfidence. Reliability diagrams on BTMRI at a 20\% label budget visualize empirical accuracy versus predicted confidence across 15 bins. Solid curves show accuracy, the dashed diagonal indicates perfect calibration, and blue bars represent the confidence distribution. SaE follows the diagonal most closely, BADGE~\cite{BADGE} is mildly underconfident, and PCB~\cite{PCB} remains strongly overconfident.}
\label{fig:reliability}
\end{figure*}

\subsubsection{Early-Round Label Efficiency}
\begin{figure}[h]
    \centering
    \includegraphics[width=\linewidth]{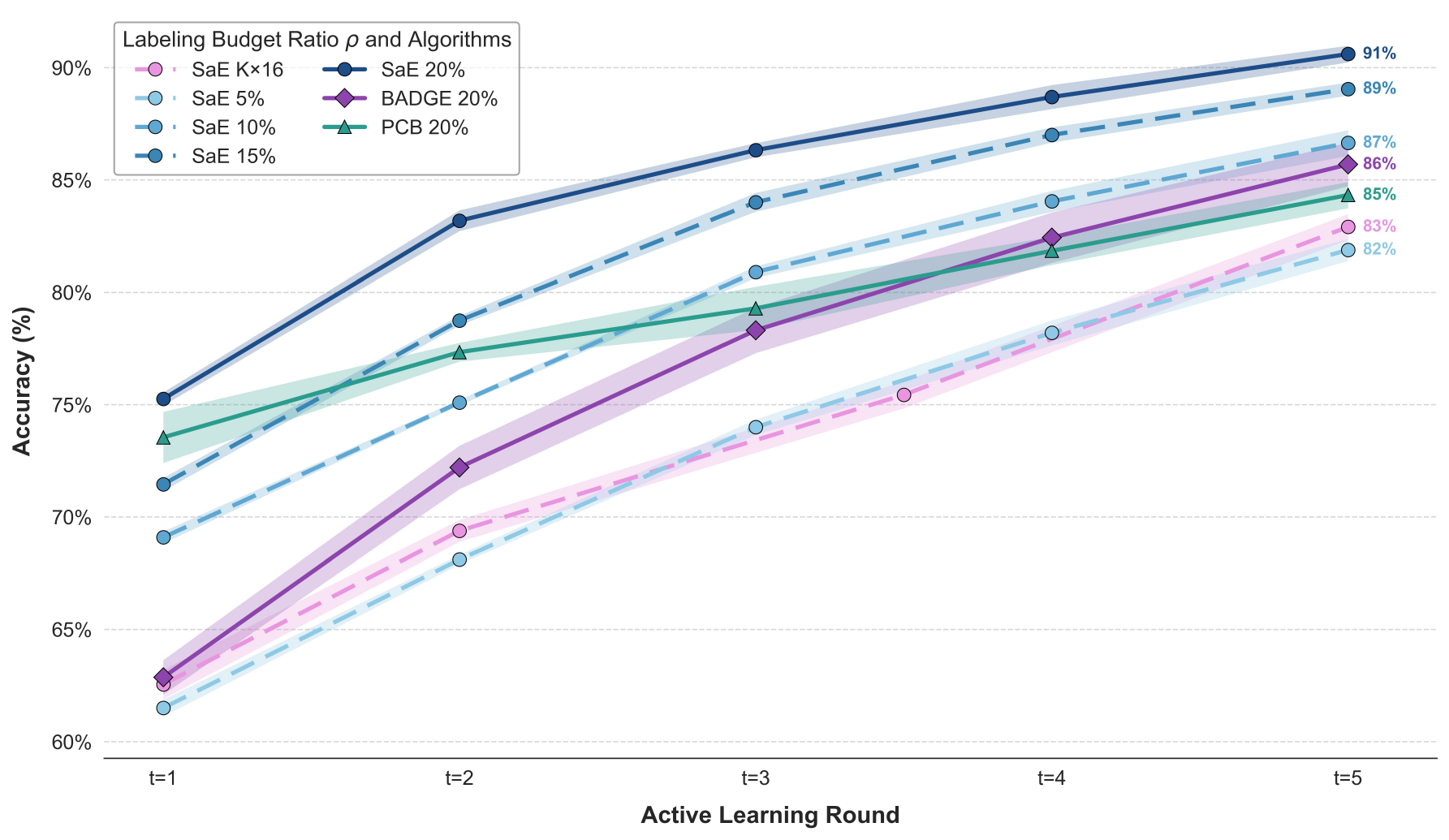}
    \caption{SaE provides superior sample efficiency, achieving higher accuracy with a 20\% budget. The mean test accuracy on CHMNIST (5 seeds, 95\% CI) is reported. SaE is shown at multiple budgets; baselines are at 20\% for comparison.}
    \label{fig:al-result}
\end{figure}

Fig.~\ref{fig:loss-trajectory} plots the training loss trajectories on BTMRI. SaE maintains the lowest and most stable loss profile from the first epoch. This contrasts sharply with BADGE, which suffers from high initial loss and significant instability, validating the cold-start problem. The VLM-based BiomedCLIP+BADGE baseline starts lower but exhibits high volatility throughout five rounds. SaE's stable and efficient optimization path provides a clear explanation for its rapid accuracy gains in early rounds.
The accuracy results in Table~\ref{tab:round_efficiency} confirm this efficiency. By $t{=}3$ (i.e., after spending 60\% of the budget), SaE reaches on average 96.7\% of its final $t{=}5$ accuracy, indicating a stable cold start. Fig.~\ref{fig:al-result} shows SaE starts stronger and gains more than BADGE~\cite{BADGE} and PCB~\cite{PCB} on CHMNIST. The confidence bands are narrow and indicate stable training. 
Under the \(K{\times}16\) setting on CHMNIST with \(K{=}8\), SaE reaches 83\% and exceeds BiomedCoOp~\cite{BiomedCoOp} by 3.95 points \((79.05\pm2.24)\).

\subsubsection{Calibration Analysis}
SaE produces well-calibrated uncertainty estimates on BTMRI, whereas VLM-based baselines remain overconfident. Calibration is evaluated on the BTMRI dataset at a 20\% labeling budget. Predictions are grouped into 15 equal-width bins on $[0,1]$, and reliability diagrams plot empirical accuracy against predicted confidence. Fig.~\ref{fig:reliability} shows that SaE attains near-perfect calibration with ECE=0.021 and NLL=0.425. As a result, probabilities from SaE can be treated as meaningful measures of evidence during AL.
PCB exhibits pronounced miscalibration with ECE=0.116 and NLL=0.757. More than 60\% of its predictions fall into the highest-confidence bin ($>0.9$), while accuracy in this bin is only about 85\%. Such overconfidence encourages the active learner to query cases that appear certain rather than informative. BADGE shows moderate miscalibration with ECE=0.036 and NLL=0.548, which is better than PCB but still yields 29\% higher NLL than SaE. 

\section{Conclusion}
We introduced SaE, an evidential AL framework that converts VLM text-image similarities into calibrated Dirichlet evidence for medical image classification. We identify that AL with pretrained VLMs faces two obstacles: a cold-start when labels are scarce and overconfident softmax outputs. The latter arises because temperature-scaled softmax treats similarity scores as deterministic, suppressing uncertainty that should be exposed during acquisition.
To address these issues, SaE equips a SEH that parameterizes a Dirichlet distribution from similarities and decomposes uncertainty, yielding clinically interpretable, actionable signals for querying. Across ten datasets with a 20\% label budget, SaE consistently outperforms existing VLM-based AL methods in both accuracy and calibration. Overall, SaE advances interpretable, label-efficient AL and strengthens the reliability of VLM-driven pipelines for clinical deployment.

\section*{Acknowledgments}
This research was funded by the Key Program of Marine Economy Development Special Foundation of Department of Natural Resources of Guangdong Province (GDNRC[2023]24).
{
    \small
    \bibliographystyle{ieeenat_fullname}
    \bibliography{main}
}

\clearpage
\setcounter{page}{1}
\maketitlesupplementary

This document provides additional details and results to support the claims made in the main paper. Unless otherwise specified, all experiments follow the same protocol and hyperparameters described in the main text.
We organize the supplementary material as follows:

\begin{itemize}
    \item \textbf{Sec.~\ref{sec:SEH-arch}} details the network architecture of the Similarity Evidence Head (SEH).
    \item \textbf{Sec.~\ref{sec:ablation}} presents comprehensive ablation studies, including the SEH loss design, sensitivity analyses of hyperparameters $\beta$ and $\epsilon$, the effect of context length $M$, and a validation of the acquisition schedule.
    \item \textbf{Sec.~\ref{sec:viz}} demonstrates the visual interpretability of SaE through qualitative Grad-CAM comparisons and uncertainty maps.
    \item \textbf{Sec.~\ref{sec:prompts}} describes the pipeline used to collect and curate PubMed-augmented prompts and illustrates it with examples on BTMRI.
\end{itemize}

\section{Similarity Evidence Head Architecture}
\label{sec:SEH-arch}
\begin{figure}
    \centering
    \includegraphics[width=1\linewidth]{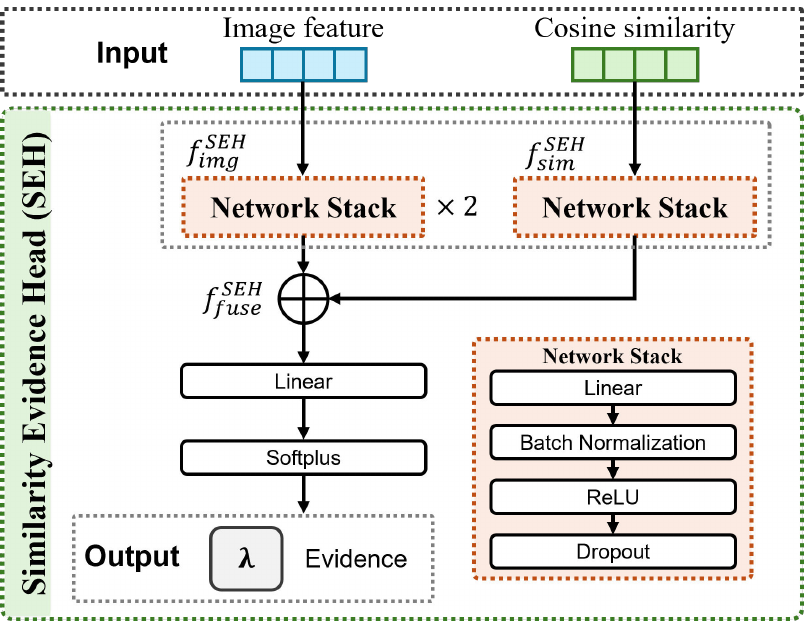}
    \caption{Architecture of the Similarity Evidence Head (SEH). SEH employs a dual-branch design to process image features $\mathbf{x}$ and similarity scores $\mathbf{s}$. The image branch consists of two stacked MLP blocks (Linear-BN-ReLU-Dropout) to extract deep semantic cues, while the similarity branch uses a single MLP block. These features are concatenated and fused via a final linear layer followed by a Softplus activation to ensure the output evidence strength $\lambda$ is strictly positive.}
    \label{fig:seh-arch}
\end{figure}

Fig.~\ref{fig:seh-arch} illustrates the SEH, a lightweight dual-branch network designed to map frozen VLM outputs to a scalar evidence strength $\lambda$. The architecture comprises an image branch that processes high-dimensional embeddings $\mathbf{x}$ through two stacked blocks to extract deep semantic cues and a similarity branch that encodes the raw similarity vector $\mathbf{s}$ via a single block. These feature streams are concatenated and fused by a final linear projection followed by a Softplus activation, strictly enforcing the positivity constraint ($\lambda > 0$) required for Dirichlet parameterization.

\section{Ablation Analysis}
\label{sec:ablation}

\subsection{Ablation on SEH Loss Design}
\label{sec:ablation_loss}

We validate the design of our Similarity Evidence Head (SEH) loss function by decomposing it into two distinct aspects: the information sources (loss components) and the mathematical formulation (regression targets). All experiments are conducted on the BTMRI dataset~\cite{BTMRI} at a 20\% budget.

\paragraph{Impact of Loss Components.}
The SEH is designed to act as a calibration bridge, connecting empirical classification difficulty ($l_{\text{cls}}$) with the VLM's intrinsic uncertainty ($H[\mathbf{p}]$). To isolate the contribution of each signal, we explicitly define the two components of the total SEH loss:
\begin{equation}
    \mathcal{L}_{\text{SEH}} = \underbrace{\operatorname{MSE}\!\left(\frac{1}{\lambda + \epsilon}, \, l_{\text{cls}}\right)}_{\mathcal{L}_{\text{diff}}} + \beta \underbrace{\operatorname{MSE}\!\left(\lambda, \, \frac{1}{H[\mathbf{p}] + \epsilon}\right)}_{\mathcal{L}_{\text{ent}}}.
\end{equation}
Here, $\mathcal{L}_{\text{diff}}$ forces the evidence strength to reflect ground-truth difficulty, while $\mathcal{L}_{\text{ent}}$ ensures consistency with the VLM's prior knowledge. We verify the necessity of this dual-objective design by training SaE with each component individually.
Table~\ref{tab:loss_components} summarizes the results. Using only entropy ($\mathcal{L}_{\text{ent}}$) yields the worst performance among the variants we test (92.15\%), suggesting that merely mimicking the VLM's existing confidence fails to correct its inherent overconfidence. Using only difficulty ($\mathcal{L}_{\text{diff}}$) improves accuracy to 92.90\% but lacks the distributional prior from the pre-trained VLM. The full dual objective ($\mathcal{L}_{\text{SEH}}$) achieves the highest accuracy of 93.46\%. This confirms that fusing empirical errors with semantic priors is crucial for robust evidence estimation.

\begin{table}[h]
    \centering
    \caption{Ablation of SEH loss components on BTMRI~\cite{BTMRI}. $\mathcal{L}_{\text{diff}}$ is the difficulty matching term, and $\mathcal{L}_{\text{ent}}$ is the entropy consistency term. The combination yields the best performance.}
    \label{tab:loss_components}
    \setlength{\tabcolsep}{10pt}
    \resizebox{1\linewidth}{!}{
        \begin{tabular}{l c c c}
            \toprule
            \textbf{Method} & \textbf{Loss Formulation} & \textbf{Acc (\%)} & \textbf{NLL} \\
            \midrule
            Entropy Only & $\mathcal{L}_\text{SEH} = \mathcal{L}_{\text{ent}}$ & 92.15 & 0.468 \\
            Difficulty Only & $\mathcal{L}_\text{SEH} = \mathcal{L}_{\text{diff}}$ & 92.90 & 0.441 \\
            \textbf{SaE (Dual)} & $\boldsymbol{\mathcal{L}_\text{SEH} = \mathcal{L}_{\text{diff}} + \beta \mathcal{L}_{\text{ent}}}$ & \textbf{93.46} & \textbf{0.425} \\
            \bottomrule
        \end{tabular}
    }
\end{table}

\paragraph{Impact of Regression Forms.}
We next investigate the mathematical form of the regression target. Our proposed method uses an inverse form ($\lambda^{-1}$), which naturally maps high uncertainty to low evidence values close to zero. We compare this against a standard logarithmic form ($-\log \lambda$), often used in uncertainty quantification to regress variance parameters.
Table~\ref{tab:loss_forms} compares these two variants. The log form achieves a competitive accuracy of 93.35\%. However, our proposed inverse form slightly outperforms it (93.46\%) and provides better calibration (lower NLL). We hypothesize that the inverse form provides stronger gradient signals for hard samples where $\lambda$ approaches zero. We therefore adopt the inverse form as our default design.

\begin{table}[h]
    \centering
    \caption{Comparison of regression target forms on BTMRI~\cite{BTMRI}. The inverse form (regressing $1/\lambda$) slightly outperforms the log form (regressing $-\log \lambda$) in both accuracy and calibration.}
    \label{tab:loss_forms}
    \setlength{\tabcolsep}{12pt}
    \resizebox{0.85\linewidth}{!}{
        \begin{tabular}{l c c c}
            \toprule
            \textbf{Regression Variant} & \textbf{Target Transform} & \textbf{Acc (\%)} & \textbf{NLL} \\
            \midrule
            Log-form & $-\log(\lambda)$ & 93.35 & 0.432 \\
            \textbf{Inverse-form (Ours)} & $(\lambda+\epsilon)^{-1}$ & \textbf{93.46} & \textbf{0.425} \\
            \bottomrule
        \end{tabular}
    }
\end{table}

\subsection{Effect of $\beta$}

The hyperparameter $\beta$ in Eq.~\ref{eq:seh-loss} controls the trade-off between fitting the empirical classification difficulty and aligning with the VLM's intrinsic entropy. We vary $\beta$ in $\{0.1, 0.3, 0.5, 0.7, 1.0\}$ while fixing $\epsilon=10^{-3}$, and measure Top-1 accuracy, NLL, and ECE at a $20\%$ budget on BTMRI~\cite{BTMRI} and RETINA~\cite{RETINA1,RETINA2}. Results are shown in Table~\ref{tab:beta-ablation}. We observe that performance is stable across a broad range of $\beta$. In practice, we set $\beta = 0.5$ for all datasets to strike a balance between the two learning objectives.

\begin{table}[h]
    \centering
    \small
    \caption{Sensitivity of SaE to the loss weight $\beta$. Performance is robust across a wide range of $\beta$, with the optimal trade-off consistently observed at $\beta=0.5$. Extreme values (0.1 or 1.0) tend to degrade both calibration (NLL/ECE) and accuracy.}
    \label{tab:beta-ablation}
    \setlength{\tabcolsep}{6pt}
    \resizebox{\linewidth}{!}{
        \begin{tabular}{l|c|ccccc}
            \toprule
            \textbf{Dataset} & \textbf{Metric} & \textbf{$\beta=0.1$} & \textbf{$\beta=0.3$} & \textbf{$\beta=0.5$} & \textbf{$\beta=0.7$} & \textbf{$\beta=1.0$} \\
            \midrule
            \multirow{3}{*}{BTMRI}
            & Acc (\%) & 92.82 & 93.18 & \textbf{93.46} & 93.24 & 92.95 \\
            & NLL      & 0.452 & 0.435 & \textbf{0.425} & 0.431 & 0.448 \\
            & ECE      & 0.029 & 0.025 & \textbf{0.021} & 0.024 & 0.027 \\
            \midrule
            \multirow{3}{*}{RETINA}
            & Acc (\%) & 73.54 & 74.65 & \textbf{75.22} & 74.92 & 74.18 \\
            & NLL      & 0.535 & 0.508 & \textbf{0.492} & 0.501 & 0.519 \\
            & ECE      & 0.047 & 0.041 & \textbf{0.039} & 0.040 & 0.044 \\
            \bottomrule
        \end{tabular}
    }
\end{table}

\subsection{Effect of $\epsilon$}
We next fix $\beta = 0.5$ and evaluate the impact of the numerical stability constant $\epsilon$ in the SEH loss on the BTMRI dataset. We vary $\epsilon \in \{10^{-4}, 5\times10^{-4}, 10^{-3}, 5\times10^{-3}, 10^{-2}\}$ and report the Top-1 accuracy at a $20\%$ budget. As shown in Table~\ref{tab:eps-ablation}, SaE maintains stable performance for $\epsilon$ between $10^{-4}$ and $10^{-2}$. We therefore fix $\epsilon=10^{-3}$ in all experiments to ensure numerical stability without compromising accuracy.

\begin{table}[h]
    \centering
    \small
    \caption{Sensitivity of SaE to the numerical stability parameter $\epsilon$ on BTMRI. SaE is largely insensitive to $\epsilon$ within a reasonable range ($10^{-4}$ to $10^{-2}$), with peak performance at the default setting.}
    \label{tab:eps-ablation}
    \setlength{\tabcolsep}{10pt}
    \resizebox{\linewidth}{!}{
        \begin{tabular}{lccccc}
            \toprule
            $\epsilon$ & $1\times10^{-4}$ & $5\times10^{-4}$ & $1\times10^{-3}$ & $5\times10^{-3}$ & $1\times10^{-2}$ \\
            \midrule
            Acc (\%) & 93.41 & 93.44 & \textbf{93.46} & 93.38 & 93.25 \\
            \bottomrule
        \end{tabular}
    }
\end{table}

\begin{figure*}[t]
    \centering
    \includegraphics[width=1\linewidth]{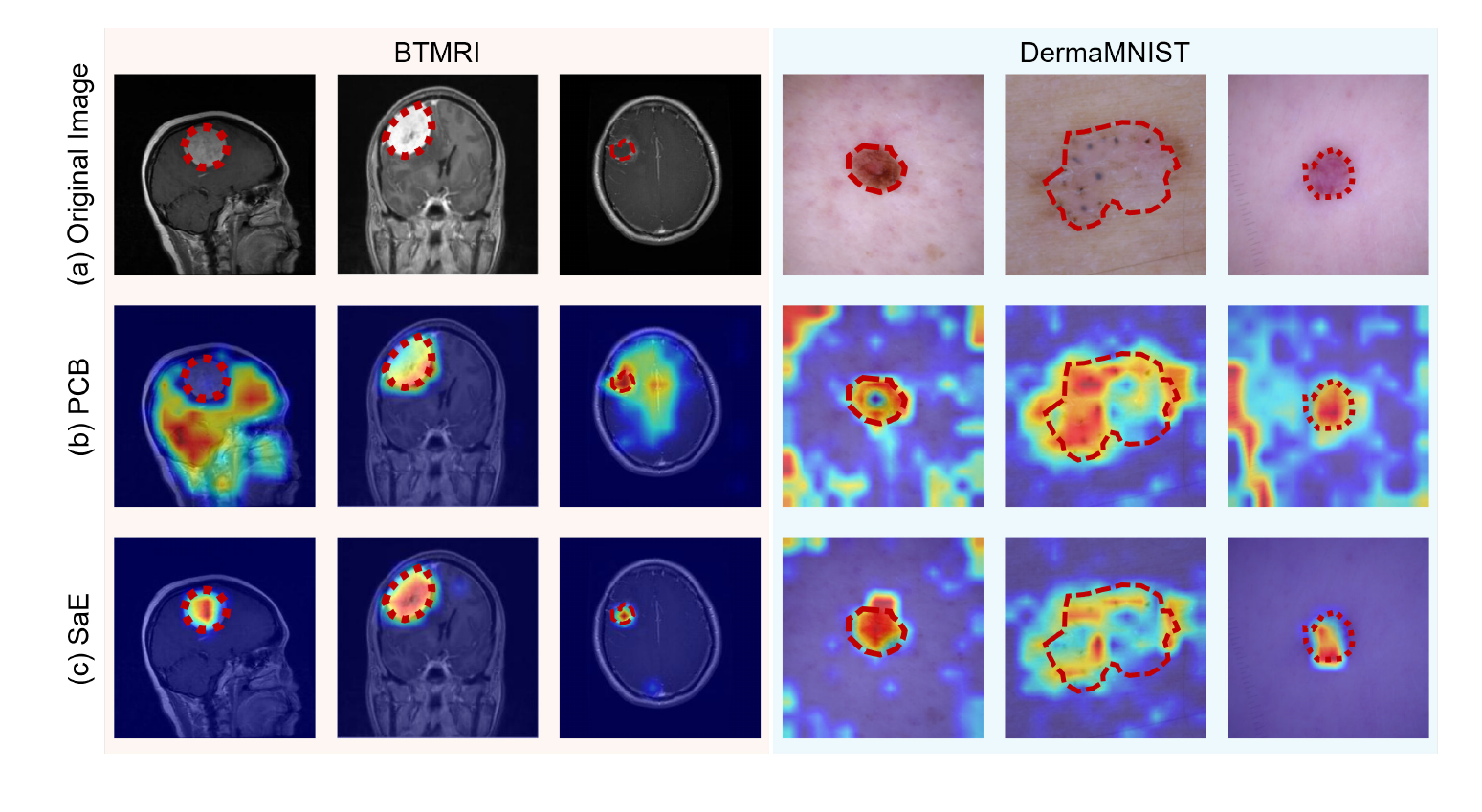}
    \caption{Visual interpretability comparison on BTMRI~\cite{BTMRI} and DermaMNIST~\cite{DermaMNIST1,DermaMNIST2}.
    We visualize Grad-CAM~\cite{Grad-Cam} activation maps for the PCB~\cite{PCB} and our SaE. 
    The red dashed contours indicate the ground-truth lesion regions. (a) Original input images. 
    (b) PCB attention is often scattered, focusing on irrelevant background regions or failing to cover the entire lesion. (c) SaE generates highly focused and accurate attention maps that align closely with the pathological regions, confirming that our evidence-calibrated strategy successfully localizes clinical features.}
    \label{fig:viz}
\end{figure*}

\subsection{Effect of Context Length}
The length of the learnable context vectors $M$ acts as a critical hyperparameter governing the trade-off between adaptation capacity and overfitting risk. We investigate the sensitivity of SaE to this parameter by varying $M \in \{4, 16, 32, 64\}$ on DermaMNIST~\cite{DermaMNIST1, DermaMNIST2} and BTMRI~\cite{BTMRI} at a fixed budget ($\rho=0.2$). Table~\ref{tab:context_length} summarizes the results. We observe that setting $M=16$ consistently yields the optimal performance across both datasets. Shorter contexts ($M=4$) appear insufficient to capture the necessary domain-specific semantics for fine-grained medical classification. Conversely, increasing the context length to 32 or 64 leads to a noticeable performance degradation. This drop likely stems from the model overfitting to the small labeled set available in active learning. Consequently, we adopt $M=16$ as the default setting to ensure robust few-shot adaptation without overfitting.

\begin{table}[ht]
    \centering
    \caption{Impact of context length $M$ on AL performance ($\rho=0.2$). A moderate length ($M=16$) achieves the best accuracy by balancing semantic capacity and overfitting. Performance drops at larger lengths ($M=32, 64$) due to overfitting on limited active learning data.}
    \label{tab:context_length}
    \setlength{\tabcolsep}{8pt}
    \resizebox{\linewidth}{!}{
        \begin{tabular}{l c c c c}
            \toprule
            \textbf{Dataset} & \textbf{$M=4$} & \textbf{$M=16$} & \textbf{$M=32$} & \textbf{$M=64$} \\
            \midrule
            DermaMNIST & 78.86 & \textbf{80.21} & 79.57 & 77.26 \\
            BTMRI      & 92.22 & \textbf{93.46} & 92.93 & 91.56 \\
            \bottomrule
        \end{tabular}
    }
\end{table}

\subsection{Effect of Acquisition Schedule}
\label{sec:ablation_schedule}
The dynamic interaction between vacuity and dissonance is critical for maximizing AL efficiency across different phases. We validate our proposed schedule by comparing it against three static baselines on the BTMRI dataset~\cite{BTMRI}. Specifically, we evaluate: (i) a vacuity-only strategy where $w_v(t)= 1$ and $w_d(t)= 0$; (ii) a dissonance-only strategy where $w_v(t)= 0$ and $w_d(t)= 1$; and (iii) a static balanced strategy where $w_v(t) = w_d(t) = 0.5$ throughout all rounds.
Table~\ref{tab:schedule_ablation} summarizes the performance differences. The dissonance-only strategy yields the poorest results (89.12\%). This significant drop confirms that prioritizing ambiguous boundary cases before the model establishes a solid knowledge base leads to severe cold-start failure. The Vacuity-only strategy performs competitively in early stages but eventually plateaus (92.55\%). This suggests that relying solely on exploration fails to refine decision boundaries in later rounds. The static balanced approach improves over single-factor methods but still lags behind our dynamic schedule. These results strongly support our explore-then-refine hypothesis, where the model benefits from prioritizing high-vacuity samples early and high-dissonance samples later.

\begin{table}[h]
    \centering
    \caption{Ablation of acquisition schedules on BTMRI~\cite{BTMRI}. The dynamic schedule outperforms all static variants. Dissonance-only fails due to cold-start instability, while Vacuity-only lacks late-stage refinement. Our dynamic strategy optimally bridges these two needs.}
    \label{tab:schedule_ablation}
    \setlength{\tabcolsep}{8pt}
    \resizebox{0.95\linewidth}{!}{
        \begin{tabular}{l c c c c}
            \toprule
            \textbf{Schedule Strategy} & \textbf{$w_v(t)$} & \textbf{$w_d(t)$} & \textbf{Acc (\%)} & \textbf{NLL} \\
            \midrule
            Dissonance-only & $= 0$ & $= 1$ & 89.12 & 0.584 \\
            Vacuity-only & $= 1$ & $= 0$ & 92.55 & 0.460 \\
            Static balanced & $= 0.5$ & $= 0.5$ & 92.85 & 0.445 \\
            \textbf{SaE (Dynamic)} & \boldmath$1 - \frac{t-1}{T-1}$ & \boldmath$\frac{t-1}{T-1}$ & \textbf{93.46} & \textbf{0.425} \\
            \bottomrule
        \end{tabular}
    }
\end{table}

\section{Visual Interpretability}
\label{sec:viz}

To qualitatively validate the source of our model's performance and calibration, we employ Grad-CAM~\cite{Grad-Cam} to visualize the regions that contribute most to the predictions. Fig.~\ref{fig:viz} compares the activation maps of our SaE framework against the PCB~\cite{PCB} on two representative datasets: BTMRI~\cite{BTMRI} (brain tumors) and DermaMNIST~\cite{DermaMNIST1, DermaMNIST2} (skin lesions).
As illustrated in the second row of Fig.~\ref{fig:viz}, the PCB often exhibits scattered attention. It frequently focuses on irrelevant background structures (e.g., the skull boundary in MRI or healthy skin texture) rather than the pathological lesion. This behavior suggests that its high confidence scores may stem from spurious correlations rather than true clinical features, which helps explain its tendency toward overconfidence on out-of-distribution samples.
In contrast, the third row shows that SaE generates highly focused attention maps that align closely with the ground-truth lesion contours (indicated by red dashed lines). Whether for the intricate boundaries of a glioma or the pigmented structure of a melanoma, SaE consistently attends to the clinically relevant regions. This visual evidence supports our claim that the evidential calibration mechanism guides the model to leverage correct semantic features, thereby providing a transparent and trustworthy basis for its uncertainty estimates.

\section{PubMed-Augmented Prompts Collection}
\label{sec:prompts}

To bridge the semantic gap between generic VLMs and specialized medical terminology, we construct class-specific imaging prompts by augmenting each category name with descriptions mined from PubMed~\cite{pubmed}. PubMed serves as the primary source of domain knowledge, while a large language model is used only to summarize and rewrite the retrieved texts into concise prompts.
For each class $c_k$ in the dataset, we perform the following steps:
\begin{enumerate}
    \item Query PubMed~\cite{pubmed} for the top 20-50 articles that mention $c_k$ together with imaging-related terms (e.g., "MRI", "X-ray", "lesion").
    \item Export the retrieved records (titles and abstracts) and extract sentences that contain both the class name $c_k$ and at least one imaging keyword (e.g., "mass", "enhancement", "hyperintense"). The selected sentences are concatenated into a short context document $D_k$ for each class.
    \item Use Google Gemini~2.5 Pro\footnote{\url{https://deepmind.google/technologies/gemini/}} to process $D_k$ together with a structured instruction prompt that asks for 3-5 sentences describing the morphology, signal characteristics, and anatomical location of $c_k$ based on the provided PubMed context.
    \item Deduplicate the resulting candidate sentences using cosine similarity (threshold $>0.9$) and discard candidates that are overly long (more than 30 words) or contain non-standard terminology.
    \item Select approximately 10 diverse descriptions per class from the remaining pool, and have a board-certified radiologist validate the final set.
\end{enumerate}

\paragraph{Extraction Prompt Template.}
The following template is used when querying Gemini for each class name with its PubMed-derived context:

{\small\ttfamily
\setlength{\parskip}{6pt}
\noindent
"Extract 3-5 imaging feature descriptions for [class\_name]. Focus on:
(1) morphology (shape, borders), (2) signal patterns (T1/T2, enhancement),
(3) location and tissue effects."
}

\subsection{Example Prompts: BTMRI Dataset}

Below we show the resulting prompts for the four classes in the BTMRI brain tumor MRI dataset, including the normal brain category.

\noindent\textbf{Glioma Tumor} ($\delta_k = 10$):

{\small\ttfamily
\setlength{\parskip}{4pt}
\noindent "In the image, a glioma tumor often appears as an irregularly shaped mass with indistinct borders blending into the surrounding brain tissue."

\noindent "The photo shows a lesion with heterogeneous signal intensity, indicating a mixture of different tissue types within the glioma."

\noindent "A key feature in the image is a ring-like enhancement pattern after the administration of contrast agent, often seen in high-grade gliomas."

\noindent "This image displays significant swelling, known as vasogenic edema, in the brain tissue surrounding the glioma tumor."

\noindent "A glioma tumor is visible as a mass causing a 'mass effect', meaning it displaces or deforms adjacent normal brain structures."

\noindent "The tumor in the photo has infiltrated the white matter tracts, which is a characteristic feature of a glioma."

\noindent "This is an image of an intra-axial lesion, meaning the glioma tumor originates from within the brain parenchyma itself."

\noindent "The photo depicts a mass with central necrosis or cystic components, appearing as a dark area within the tumor."

\noindent "On this scan, the glioma appears as a poorly-defined, infiltrative lesion within the cerebral hemisphere."

\noindent "The image shows a brain with a glioma, characterized by its invasive nature and lack of a clear capsule."
}

\normalsize\normalfont
\noindent\textbf{Meningioma Tumor} ($\delta_k = 10$):

{\small\ttfamily
\setlength{\parskip}{4pt}

\noindent "The image displays a meningioma as a well-defined, distinctly bordered mass located outside the brain tissue (extra-axial)."

\noindent "A characteristic feature in the photo is the tumor's broad attachment to the dura, the outer lining of the brain."

\noindent "This is a photo of a meningioma showing intense and uniform enhancement after contrast injection, appearing as a brightly lit mass."

\noindent "In the image, a 'dural tail sign' is visible, which looks like a tail of enhancement extending from the tumor along the dura."

\noindent "The photo shows a meningioma causing compression and displacement of the adjacent brain without invading it."

\noindent "A meningioma often appears as a rounded or lobulated mass located on the surface of the brain."

\noindent "The tumor in this image is isointense to gray matter on T1-weighted images, meaning it has a similar shade before contrast."

\noindent "This image shows a meningioma that may contain calcifications, which appear as very dark or bright spots depending on the imaging sequence."

\noindent "The photo depicts a smoothly marginated mass consistent with a meningioma, pressing on the cerebral cortex."

\noindent "Visible in the scan is a dural-based mass, a classic presentation of a meningioma tumor."
}

\normalsize\normalfont
\noindent\textbf{Pituitary Tumor} ($\delta_k = 10$):

{\small\ttfamily
\setlength{\parskip}{4pt}
\noindent "The image shows a well-defined, rounded mass located within the sella turcica, the bony cavity at the base of the skull where the pituitary gland sits."

\noindent "This photo depicts an expansion of the sella turcica caused by a pituitary tumor."

\noindent "A pituitary adenoma is visible in the scan, often appearing as a lesion with a signal intensity different from the normal pituitary gland."

\noindent "In the image, the pituitary tumor can be seen extending upwards (suprasellar extension), potentially compressing the optic chiasm."

\noindent "This is a photo of a pituitary macroadenoma, defined as a tumor larger than 10mm in diameter, filling the pituitary fossa."

\noindent "The image shows a pituitary tumor that enhances less avidly and more slowly than the surrounding normal pituitary gland after contrast."

\noindent "On this sagittal view, a mass is clearly visible within the pituitary fossa, characteristic of a pituitary tumor."

\noindent "The photo shows a lesion at the base of the brain consistent with a pituitary tumor, which can sometimes be cystic or hemorrhagic."

\noindent "This scan reveals a pituitary tumor causing thinning and remodeling of the bone of the sella turcica."

\noindent "The image displays a distinct mass in the sellar region, which is the typical location for a pituitary tumor."
}

\normalsize\normalfont
\noindent\textbf{Normal Brain} ($\delta_k = 10$):

{\small\ttfamily
\setlength{\parskip}{4pt}

\noindent "The image displays a normal brain with clear differentiation between the gray matter on the outside and the white matter on the inside."

\noindent "In this photo, the brain structures are symmetrical on both the left and right sides, with no evidence of displacement."

\noindent "A normal brain scan shows the ventricles, the fluid-filled spaces, as well-defined structures of normal size and shape."

\noindent "The image shows no signs of abnormal masses, lesions, or growths within the brain tissue."

\noindent "This is a photo of a healthy brain where there is no abnormal enhancement after the administration of a contrast agent."
        
\noindent "In the image, the sulci and gyri, which are the grooves and folds of the brain cortex, appear normal and are not effaced."

\noindent "The scan shows a brain with no evidence of swelling (edema), hemorrhage, or fluid collections."

\noindent "This photo displays normal brain anatomy with all major structures, like the cerebrum, cerebellum, and brainstem, appearing unremarkable."
        
\noindent "A normal brain image is characterized by the absence of any pathological findings such as tumors, infarcts, or inflammation."

\noindent "The image shows a brain with normal signal intensity throughout, without any bright or dark spots that would suggest pathology."
}

\end{document}